\documentclass[9pt, conference]{IEEEtran}
\IEEEoverridecommandlockouts

\usepackage{cite}
\usepackage{amsmath,amssymb,amsfonts}
\usepackage{mathrsfs}
\usepackage{newtxtext}
\usepackage{newtxmath}
\usepackage{algorithmic}
\usepackage{graphicx}
\usepackage{textcomp}
\usepackage{hyperref}
\usepackage{stfloats}
\usepackage{algorithm}
\usepackage{caption}
\usepackage{pifont} 
\usepackage{pgfmath}

\usepackage{siunitx}
\usepackage{array}
\usepackage{pgf}

\usepackage{setspace}
\usepackage{xcolor} 
\usepackage{colortbl}
\usepackage{adjustbox}
\usepackage{enumitem}
\usepackage{cleveref}
\usepackage{multirow}
\usepackage{longtable}
\usepackage{rotating}
\usepackage{booktabs}
\usepackage{xurl}
\usepackage{makecell}
\usepackage[most]{tcolorbox} 
\tcbuselibrary{skins, breakable}

\usepackage[compact]{titlesec}

\titlespacing\section{0pt}{0.3\baselineskip}{0.2\baselineskip}
\titlespacing\subsection{0pt}{0.2\baselineskip}{0.1\baselineskip}
\titlespacing\subsubsection{0pt}{0.15\baselineskip}{0.1\baselineskip}

\definecolor{headercolor}{rgb}{0.2,0.4,0.6}
\definecolor{rowcolor}{gray}{0.9}
\definecolor{bordercolor}{rgb}{0.1,0.2,0.4}
\def\BibTeX{{\rm B\kern-.05em{\sc i\kern-.025em b}\kern-.08em
    T\kern-.1667em\lower.7ex\hbox{E}\kern-.125emX}}

\newcommand{\ftcross}{\textcolor{red}{\ding{55}}}
\newcommand{\ftcheck}{\textcolor{teal}{\ding{51}}}

\tcbuselibrary{listings,skins,breakable}

\lstdefinestyle{pennycode}{
    basicstyle      = \ttfamily\scriptsize,  
    keywordstyle    = \color{blue},
    commentstyle    = \color{green!60!black},
    stringstyle     = \color{red},
    numbers         = left,
    numberstyle     = \tiny\color{gray},
    stepnumber      = 1,
    numbersep       = 5pt,
    backgroundcolor = \color{white},
    frame           = single,
    rulecolor       = \color{black!30},
    tabsize         = 4,
    captionpos      = b,
    breaklines      = true,
    breakatwhitespace = true,
    showstringspaces  = false
}
\usepackage{fancyhdr,lipsum}
\fancypagestyle{firstpage}{
  \fancyhf{}
  \fancyhead[C]{To appear at the IEEE International Conference on Quantum Artificial Intelligence (QAI), Naples, Italy, November 2025.}
  \fancyfoot[C]{\thepage}
}

\begin{document}

\title{QHackBench: Benchmarking Large Language Models for Quantum Code Generation Using PennyLane Hackathon Challenges\\
\vspace{-8pt}
}

\author{
\IEEEauthorblockN{\authorsize Abdul Basit\textsuperscript{1}, Minghao Shao\textsuperscript{1}, Muhammad Haider Asif\textsuperscript{1,2}, Nouhaila Innan\textsuperscript{1,2}, Muhammad Kashif\textsuperscript{1,2}, Alberto Marchisio\textsuperscript{1,2}, \\Muhammad Shafique\textsuperscript{1,2}}
\IEEEauthorblockA{\textit{\textsuperscript{1}eBRAIN Lab, Division of Engineering} \textit{New York University (NYU) Abu Dhabi}, Abu Dhabi, UAE\\
\textit{\textsuperscript{2}Center for Quantum and Topological Systems (CQTS), NYUAD Research Institute, New York University Abu Dhabi}, UAE\\
\{abdul.basit, shao.minghao, ma8183, nouhaila.innan, muhammadkashif, alberto.marchisio, muhammad.shafique\}@nyu.edu 
\vspace{-13pt}
}}

\maketitle
\thispagestyle{empty}
\thispagestyle{firstpage}


\begin{abstract}
Recent advances in Large Language Models (LLMs) have demonstrated strong potential in code generation, yet their effectiveness in quantum computing remains underexplored. This paper benchmarks LLMs for PennyLane-based quantum code generation using real-world challenges from the Quantum Hackathon (QHack). We introduce \textit{QHackBench}, a novel benchmark dataset derived from QHack competitions, and evaluate model performance under vanilla prompting and Retrieval-Augmented Generation (RAG). Our structured evaluation framework assesses functional correctness, syntactic validity, and execution success across varying challenge difficulties.
Results indicate that RAG-enhanced models, supplemented with an augmented PennyLane dataset, approximately generate similar results as the standard prompting, particularly in complex quantum algorithms. Additionally, we introduce a multi-agent evaluation pipeline that iteratively refines incorrect solutions, further enhancing execution success rates. To foster further research, we commit to publicly releasing \textit{QHackBench}, along with our evaluation framework and experimental results, enabling continued advancements in AI-assisted quantum programming.
\end{abstract}


\begin{IEEEkeywords}
Quantum Computing, PennyLane, Large Language Models, Code Generation, Benchmarking, Retrieval-Augmented Generation.
\end{IEEEkeywords}

\section{Introduction and Motivation}

Quantum computing has emerged as a transformative technology, offering approaches for potentially solving problems considered classically intractable~\cite {arute2019quantum,kashif:2025computational,kashif:2022_demonstrating}. With the increasing adoption of hybrid quantum-classical computing, tools such as PennyLane have gained prominence \cite{Bergholm2018PennyLane}, facilitating Quantum Machine Learning (QML) and variational circuit optimization \cite{biamonte2017quantum, zaman2023survey, innan2025optimizing,innan2025qnn,innan2025next}. However, programming in quantum frameworks remains challenging due to the complexity of quantum circuits and domain-specific syntax. 

Large Language Models (LLMs)~\cite{shao2024survey} such as GPT-4 have demonstrated remarkable performance in classical code generation and completion~\cite{Dupuis2024QiskitAssistant}. However, their capabilities in quantum programming, particularly within the PennyLane framework, remain largely unexamined. Existing benchmarks, such as Qiskit HumanEval, primarily evaluate LLM performance for IBM's Qiskit framework~\cite{Vishwakarma2024QiskitHumanEval}, leaving a significant gap in assessing LLM effectiveness for PennyLane-based quantum programming. This gap is critical, as PennyLane provides unique features such as automatic differentiation for QML and seamless hardware integration~\cite{Bergholm2018PennyLane}, necessitating a dedicated benchmark.

To address this issue, we introduce a novel benchmarking approach that leverages \textit{Quantum Hackathon Challenges (QHack 2023–2024)} as an evaluation dataset, integrated with our baseline evaluation framework. The framework incorporates both vanilla prompting and Retrieval-Augmented Generation (RAG), as well as a more complex agentic setup, to account for diverse usage scenarios aligned with modern LLM-based systems. It further employs a diverse set of model families to enhance flexibility and performance. Within this setting, we systematically analyze LLM-generated quantum code and compare a wide range of configurations to establish a comprehensive baseline on our proposed benchmark. Our study shows that while RAG improves accuracy in specific cases, such as reducing hallucinated API calls and refining circuit structures, current foundation models still face significant limitations in generating reliable quantum code.


\subsection{Motivational Case Study}

In our evaluation using the o3-mini model, the generated code was assessed on a challenge-by-challenge basis as either correct or incorrect. In this study, we report only binary outcomes—each challenge is either solved correctly or not. Our experiments show that when the model’s queries were supplemented with relevant PennyLane documentation via RAG, the number of challenges solved were approximately the same as the baseline approach. Furthermore, integrating a multi-agent refinement pipeline further improved the execution success rate significantly. These findings underscore the importance of iterative debugging and the inclusion of domain-specific documentation to enhance the reliability of LLM-assisted quantum programming.



\begin{figure}[t!]
    \centering
    \includegraphics[width=\linewidth]{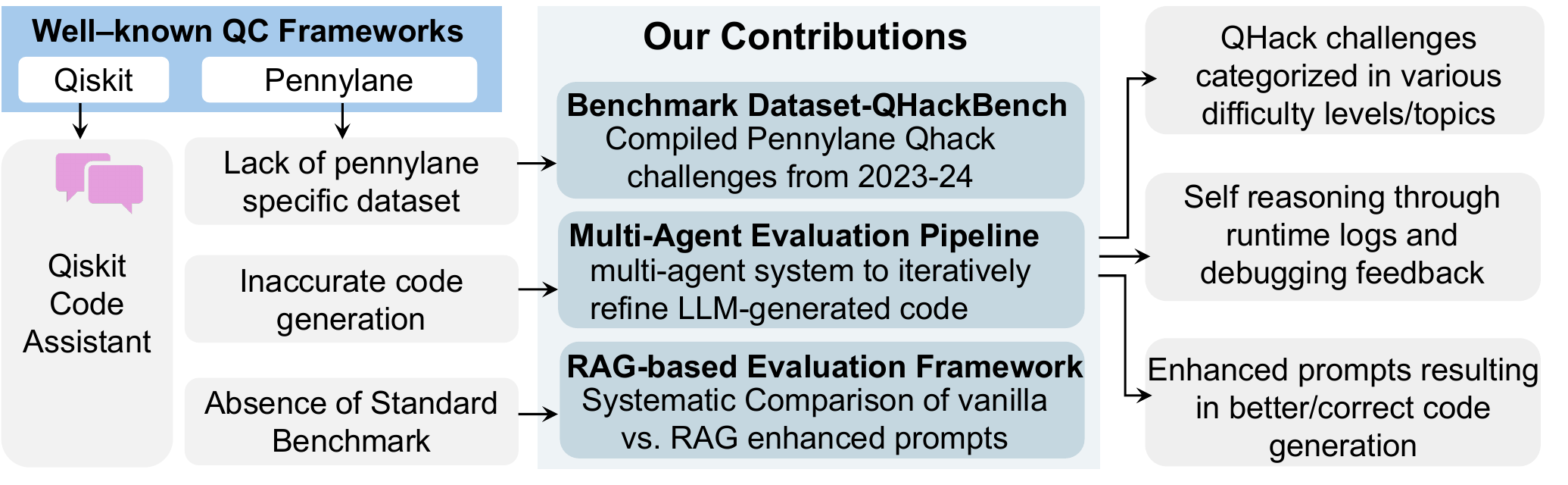}
    \vspace{-15pt}
    \caption{An overview of our contributions.}
    \label{fig:contribution}
\end{figure}
\subsection{Research Challenges} This work addresses several key challenges in LLM-based quantum code generation:
\begin{itemize}[leftmargin=*]
    \item \textbf{Lack of PennyLane-Specific Training Data:} Current LLMs are trained on limited PennyLane-related code, often leading to incorrect function calls.
    \item \textbf{Inaccurate Code Generation:} Without additional context, LLMs struggle with function signatures, quantum execution flow, and circuit correctness.
    \item \textbf{Need for a Standardized Benchmark:} Unlike Qiskit, PennyLane lacks a dedicated benchmark for evaluating LLM-assisted code generation.
\end{itemize}

\subsection{Novel Contributions}
To overcome the above challenges, we make the following key contributions, which are also summarized in Fig.~\ref{fig:contribution}.
\begin{itemize}[leftmargin=*]
    \item \textbf{Benchmark Dataset:} We introduce \textit{QHackBench}, the first curated dataset of PennyLane coding challenges from QHack 2023-2024, specifically designed to evaluate LLM performance in quantum programming.
    \item \textbf{Baseline Evaluation Agent Framework:} We propose a structured evaluation framework that iteratively refines LLM-generated quantum code through automated execution, self-reasoning, and debugging feedback, while incorporating modern agent technologies such as RAG and multi-agent systems.
    \item \textbf{Comprehensive baseline Evaluation:} We systematically compare LLMs with various configurations, assessing the impact of different features on quantum code correctness and execution success.
\end{itemize}

\textbf{Paper Organization:} \Cref{sec:background} discusses the basic concepts to understand the rest of the paper and presents the related work. \Cref{sec:methodology} introduces our proposed QHackBench methodology. \Cref{sec:exp_setup} discusses the experiment settings. \Cref{sec:results} reports the experimental results. \Cref{sec:conclusion} concludes the paper.
\section{Background and Related Work}
\label{sec:background}

\subsection{Large Language Models (LLMs) for Code Generation}
LLMs have demonstrated remarkable capabilities in code generation, as evidenced by models such as Codex~\cite{Chen2021Evaluating}, StarCoder~\cite{Li2023StarCoder}, and IBM's Granite models~\cite{Dupuis2024QiskitAssistant}. These models are trained on extensive programming datasets and benchmarked using standardized datasets like HumanEval~\cite{Chen2021Evaluating} and MBPP~\cite{Austin2021Program}. However, their effectiveness in niche domains such as quantum computing is limited due to the scarcity of domain-specific training data~\cite{Watson2022QHack}.

Prior research has explored the application of LLMs to quantum programming, primarily focusing on IBM's Qiskit framework. The Qiskit Code Assistant was fine-tuned to improve accuracy in quantum code generation tasks~\cite{GarciaAlmeida2024QiskitBlog}. Additionally, Vishwakarma et al. introduced Qiskit HumanEval, a benchmark designed to assess LLM-generated Qiskit code~\cite{Vishwakarma2024QiskitHumanEval}. However, despite PennyLane’s increasing adoption in QML, there has been limited research on evaluating LLMs for PennyLane-based quantum programming~\cite{basit2025pennycoder}. Our work fills this gap by systematically benchmarking LLM performance on PennyLane tasks using real-world coding challenges.

\subsection{Quantum Programming and PennyLane}
Quantum computing frameworks facilitate various approaches to quantum programming. IBM’s Qiskit focuses on hardware-oriented circuit execution, while Google’s Cirq optimizes for near-term quantum devices~\cite{Gidney2021Cirq}. PennyLane, developed by Xanadu, is distinguished by its emphasis on quantum differentiability, making it particularly well-suited for hybrid quantum-classical computations~\cite{Bergholm2018PennyLane}. By enabling automatic differentiation of quantum circuits, PennyLane supports the optimization of variational quantum algorithms and QML models.

Despite PennyLane’s advantages, there is currently no dedicated benchmark for evaluating LLM-generated code within this framework. The lack of large-scale PennyLane datasets presents a unique challenge, necessitating alternative evaluation methods such as RAG~\cite{Kharkov2024BraketRAG}. Our work introduces \textit{QHackBench}, the first structured benchmark dataset for PennyLane-based LLM evaluation, leveraging real-world challenges from QHack 2023-2024.

\subsection{Retrieval-Augmented Generation (RAG) for Quantum Code}
RAG has emerged as an effective technique for enhancing LLMs by incorporating relevant external knowledge sources dynamically~\cite{Lewis2020RetrievalAugmented}. In classical programming, RAG has been employed to improve code completion by retrieving real-time documentation references~\cite{Kharkov2024BraketRAG}. In the quantum computing domain, AWS Braket has integrated RAG-based solutions to improve LLM-generated quantum code by providing access to up-to-date quantum-specific knowledge.

\begin{figure}[h]
    \centering
    \includegraphics[width=0.45\textwidth]{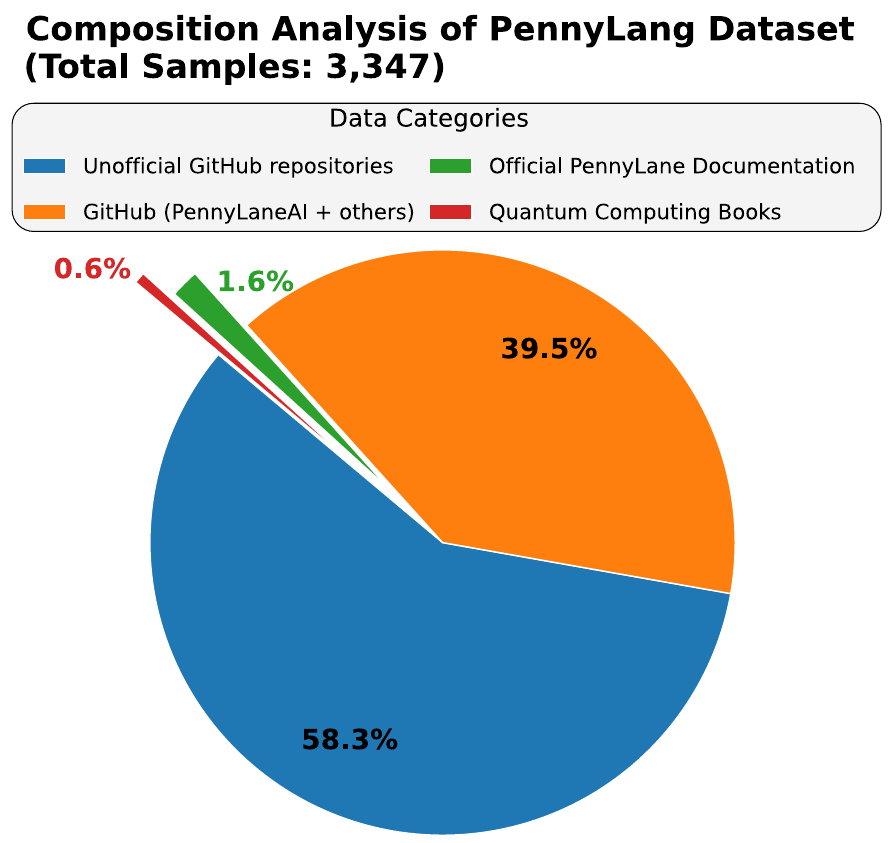}
    \caption{Composition analysis of the \textit{PennyLang} dataset (3,347 samples).}
    \label{fig:pennylang_composition}
\end{figure}

Similarly, \textit{PennyLang}, a curated dataset of PennyLane-specific code examples and quantum computing resources, serves as a structured retrieval corpus that enhances LLM prompts with relevant contextual information. By leveraging PennyLang, LLMs can access previously solved quantum problems, reducing API hallucinations and improving functional correctness in PennyLane programming~\cite{PennyLang}. The dataset consists of four primary sources: unofficial GitHub repositories, official PennyLane documentation, open-source contributions from PennyLaneAI, and quantum computing books. The composition of the dataset is shown in Fig.~\ref{fig:pennylang_composition}.

\subsection{Multi-Agent Systems for Quantum Code Generation}
Multi-agent systems have been widely explored in AI research to enhance problem-solving by enabling specialized agents to collaborate on complex tasks~\cite{Wooldridge2009MAS}. In programming, multi-agent frameworks have been employed for code synthesis and debugging, where different agents handle distinct roles such as code generation, execution validation, and iterative refinement~\cite{jin2024rgdmultillmbasedagent}. 

For quantum programming, a multi-agent approach is particularly valuable due to the inherent complexity of debugging quantum circuits, where execution errors often stem from subtle query misinterpretations or incorrect circuit logic~\cite{yang2025coastenhancingcodedebugging}.

Our experimental results demonstrate that integrating a multi-agent framework significantly improves execution success rates compared to standalone prompting or RAG-based approaches. By systematically evaluating LLM performance in a multi-agent setting, we provide the first in-depth analysis of iterative refinement in quantum code generation.


\

\section{QHackBench Methodology}
\label{sec:methodology}

QHackBench systematically evaluates LLM performance in quantum code generation through an automated benchmarking framework for QHack challenges. Our methodology integrates RAG with the PennyLang dataset to enhance model responses, benchmarking them against non-RAG baselines. The evaluation assesses correctness, execution success, and iterative improvements within a structured pipeline.

\begin{figure*}[ht]
    \centering
    \includegraphics[width=0.92\textwidth]{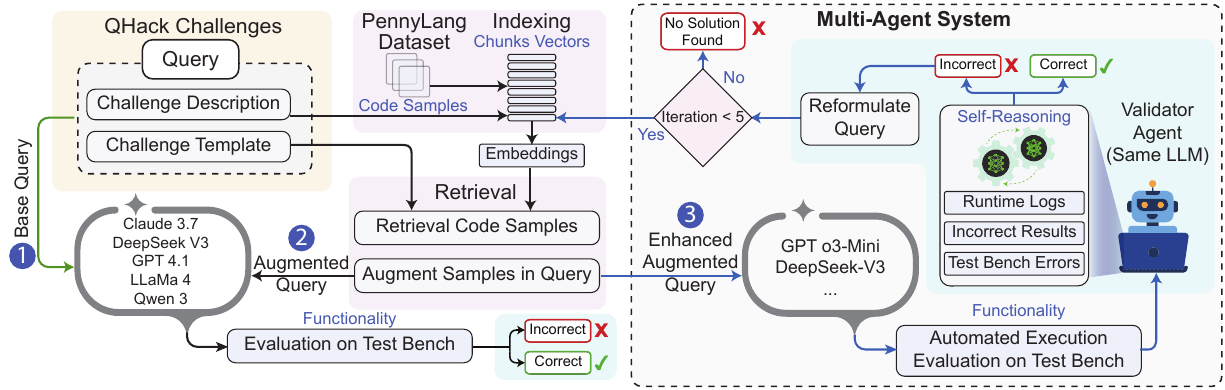}
    \vspace{-1pt}
    \caption{Overview of the \textit{QHackBench}. 
    \textcircled{\tiny 1}~\emph{Base query.} The QHack challenge description and template are provided to LLMs (GPT-4.1, Claude 3.7, ...) for initial code generation.  
    \textcircled{\tiny 2}~\emph{Retrieval-augmented query} If enabled, relevant PennyLang code snippets are retrieved, embedded, and incorporated into the query to enhance accuracy.  
    \textcircled{\tiny 2}~\emph{Multi-agent pipeline:} a validator agent, implemented with the same LLM, inspects run-time logs and hidden-test feedback, performs self-reasoning, reformulates the query, and asks the builder to regenerate code. This ensures correctness evaluation, error diagnosis, and improved execution.}
    
    \label{fig:methodology}
\end{figure*}

\subsection{Automated Evaluation Pipeline}

To ensure consistency and reproducibility in benchmarking LLMs for quantum code generation, we design a structured evaluation pipeline that systematically assesses model performance across different configurations. The overall evaluation pipeline is illustrated in Algorithm~\ref{alg:benchmark_pipeline}. The process begins with {dataset preparation}, where QHack challenge descriptions, template notebooks, and corresponding test scripts from the years 2023 and 2024 are gathered to establish a standardized benchmark. Next, we integrate RAG by retrieving relevant code snippets from the PennyLang dataset \cite{PennyLang}, providing models with enhanced contextual knowledge for more accurate code completion. To analyze the effectiveness of RAG, we evaluate multiple LLMs, including {GPT-4o mini (RAG and non-RAG)} and {GPT-o3 Mini (RAG and non-RAG)}, allowing a direct comparison between retrieval-enhanced responses and baseline prompting.

Once the solutions are generated, we employ an {automated execution system} that runs the code in an isolated environment, verifying functional validity and correctness against predefined test cases. This is followed by the {performance evaluation}, where we measure key metrics such as Pass@$k$, execution success rate, and error classification to quantify model effectiveness. Additionally, we incorporate an {iterative refinement process using a multi-agent system}, wherein an automated validation mechanism analyzes execution errors and refines incorrect solutions iteratively. By following this structured sequence, spanning code generation, retrieval augmentation, correctness verification, iterative improvement, and automated execution, our pipeline provides a robust and scalable framework for benchmarking LLMs in quantum programming, enabling precise evaluation of their capabilities in solving QHack challenges.

\begin{algorithm}[h]
\scriptsize
\caption{Automated Benchmarking Pipeline for QHack Challenges}
\label{alg:benchmark_pipeline}
\begin{algorithmic}[1]
    \REQUIRE Quantum challenge dataset $\mathcal{D}$, LLM models $\mathcal{M}$, PennyLang retrieval corpus $\mathcal{R}$
    \FOR{each challenge $(C_i, T_i) \in \mathcal{D}$}
        \STATE Retrieve relevant context $\mathcal{S}_i$ from $\mathcal{R}$ using embedding similarity.
        \FOR{each model $M_j \in \mathcal{M}$}
            \STATE Generate quantum code $\mathcal{Q}_i^j$ using RAG-assisted and non-RAG prompting.
            \STATE Execute $\mathcal{Q}_i^j$ in a controlled environment.
            \STATE Collect execution logs $\mathcal{L}_i^j$ and correctness status.
        \ENDFOR
        \STATE Perform error analysis and refine $\mathcal{Q}_i^j$ iteratively (if multi-agent mode is enabled).
    \ENDFOR
    \STATE Compute benchmark metrics: Pass@$k$, execution success rate, error classification.
\end{algorithmic}
\end{algorithm}

\subsection{Retrieval-Augmented Generation with PennyLang Dataset}

To enhance LLM-based quantum code generation, we integrate retrieval from the PennyLang dataset, a curated collection of resources relevant to PennyLane-based quantum programming. For each challenge, a similarity-based retriever selects the most relevant context from $\mathcal{R}$ using a vector database. This retrieved knowledge is appended to the model prompt, improving accuracy and reducing hallucinations. The RAG process is outlined in Algorithm~\ref{alg:rag_retrieval}.

\begin{algorithm}[h]
\scriptsize
\caption{Retrieval-Augmented Generation (RAG) for Quantum Code}
\label{alg:rag_retrieval}
\begin{algorithmic}[1]
    \REQUIRE Query input $Q$, retrieval corpus $\mathcal{R}$, embedding model $\mathcal{E}$
    \STATE Compute query embedding $e_Q = \mathcal{E}(Q)$
    \STATE Retrieve top-$k$ documents $\mathcal{S} = \arg \max_{S_i \in \mathcal{R}} \text{similarity}(e_Q, \mathcal{E}(S_i))$
    \STATE Construct prompt: $\mathcal{P} = \text{system\_prompt} + \mathcal{S} + Q$
    \STATE Generate response: $\mathcal{Q} = \text{LLM}(\mathcal{P})$
    \RETURN Generated code $\mathcal{Q}$
\end{algorithmic}
\end{algorithm}






\subsection{Multi-Agent System for Iterative Code Refinement}

To improve the reliability of LLM-generated quantum code, we employ a {multi-agent system} that iteratively refines incorrect solutions. The system consists of two main agents: 

\begin{itemize}[leftmargin=*]
    \item \textbf{Code Generation Agent} (GPT-o3-Mini, DeepSeek-V3): Generates initial quantum code solutions using either RAG or non-RAG methods.
    \item \textbf{Validation \& Correction Agent} (GPT-o3-Mini): Executes the generated code, analyzes errors from execution logs, and iteratively refines the solution through self-reasoning and debugging.
\end{itemize}

The multi-agent system follows a structured pipeline designed to iteratively improve LLM-generated quantum code by leveraging both RAG and automated debugging. The process begins with the {Code Generation Agent}, which produces an initial quantum circuit implementation based on the QHack challenge description. This generated solution is then executed within a controlled Jupyter environment, where its correctness is evaluated against predefined test cases. If execution fails, the {Validation \& Correction Agent} analyzes the runtime logs, extracting error messages, incorrect results, and test bench failures. Based on this analysis, the agent formulates a structured debugging report, identifying potential issues and suggesting modifications. The revised query, incorporating these debugging insights, is then sent back to the {Code Generation Agent} for solution refinement. If RAG is enabled, the retrieved context from the \textit{PennyLang dataset} is also included in the updated query to enhance code corrections. This process is repeated iteratively, with a maximum of five refinement attempts, ensuring that the solution is progressively improved until it successfully passes all test cases or reaches the retry limit.

\section{Experimental Setup}
\label{sec:exp_setup}

\subsection{Benchmark Dataset}
\begin{figure}[t]
    \centering
    \includegraphics[width=0.43\textwidth]{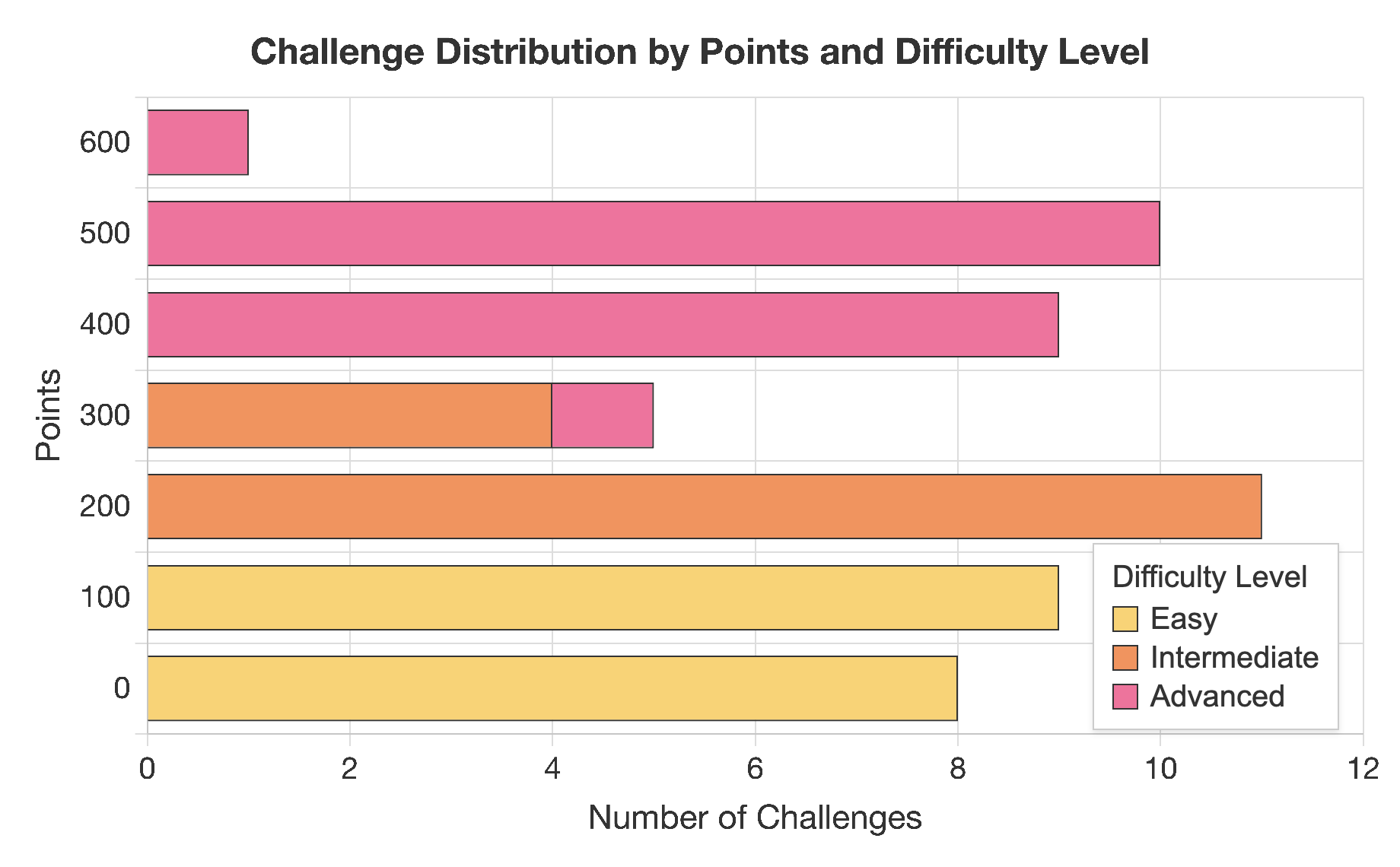}
    \caption{Difficulty distribution among all the challenges.}
    \label{fig:distribution}
\end{figure}

We curated a dataset comprising 49 unique QHack challenges from 2023-2024 \cite{XanaduAI2,XanaduAI3}.
These challenges span multiple QC domains, including quantum algorithms, QML, and quantum chemistry. The dataset is structured based on the yearly challenge formats and categorized by difficulty levels and thematic divisions.
\subsection{Structure of QHack Challenges}
QHack challenges adhere to a structured format, with problems assigned to predefined categories and difficulty levels. The dataset is categorized as follows:
\begin{itemize}[leftmargin=*]
\item \textbf{2023:} Challenges were structured within five narrative-based thematic sets: A Tale of Timbits, Bending Bennett’s Laws, Fall of Sqynet, and Office Hijinks. Each narrative contained five challenges, with a standardized point distribution ranging from 100 to 500 points. A total of 28 challenges were included, with 8 tutorial challenges assigned 0 points. The introduction of a narrative-driven format added an engaging element to problem-solving.
\item \textbf{2024:} The challenge structure evolved further with a thematic-based organization, introducing five core themes: Boson Beach, Dipole Desert, FemtoForest, Tensor Tundra, and Qutropolis. This iteration featured 21 challenges with difficulty levels spanning 100 to 600 points. Higher-difficulty problems were more prevalent in later themes, reflecting an increased focus on complex quantum problem-solving.
\end{itemize}
\begin{table}[htpb]
    \centering

   \caption{Full challenge list of QHackBench.}
\label{tab:challenge}
\begin{adjustbox}{max width=1\linewidth}

\begin{tabular}{|c|l|l|c|c|}

    \hline
    \rowcolor{gray} \textcolor{white}{\textbf{Year}} & \textcolor{white}{\textbf{Theme}} & \textcolor{white}{\textbf{Challenge}} & \textcolor{white}{\textbf{Points}} & \textcolor{white}{\textbf{Difficulty}} \\
    \hline

 \multirow{28}{*}{\rotatebox{90}{\textbf{2023} (28 Challenges)}}
    & \multirow{8}{*}{Tutorials} & Tutorial 1 (C1-E-23)  & 0 & Easy \\
    & & Tutorial 2 (C2-E-23) & 0 & Easy \\
    & & Tutorial 3 (C3-E-23) & 0 & Easy \\
    & & Tutorial 4 (C4-E-23)& 0 & Easy \\
    & & Tutorial 5 (C5-E-23) & 0 & Easy \\
    & & Tutorial 6 (C6-E-23) & 0 & Easy \\
    & & Tutorial 7 (C7-E-23) & 0 & Easy \\
    & & Tutorial 8 (C8-E-23) & 0 & Easy \\
    \cline{2-5}
    
    & \multirow{5}{*}{\makecell{A Tale of\\ Timbits}} & The Magic 5 Ball (C9-E-23) & 100 & Easy \\
    & & Cascadar (C10-I-23) & 200 & Intermediate \\
    & & A Pauli Worded Problem (C11-I-23) & 300 & Intermediate \\
    & & Entangled Robot Swarms (C12-A-23) & 400 & Advanced \\
    & & One Bit Wonder (C13-A-23) & 500 & Advanced \\
    \cline{2-5}
    
    & \multirow{5}{*}{\makecell{Bending\\ Bennett's\\ Laws}} & Ctrl-Z (C14-E-23) & 100 & Easy \\
    & & Sub Superdense Coding (C15-I-23) & 200 & Intermediate \\
    & & Secrets in Spacetime (C16-I-23)& 300 & Intermediate \\
    & & A Halfway Decent Photocopier (C17-A-23) & 400 & Advanced \\
    & & The Itch to Switch (C18-A-23) & 500 & Advanced \\
    \cline{2-5}
    
    & \multirow{5}{*}{\makecell{Fall of\\ Sqynet}} & Sqy Trotter (C19-E-23) & 100 & Easy \\
    & & Unitary Operators (C20-I-23) & 200 & Intermediate \\
    & & Don't Hit the Ground (C21-I-23)& 300 & Intermediate \\
    & & Desperate Measures (C22-A-23) & 400 & Advanced \\
    & & Ising Uprising (C23-A-23) & 500 & Advanced \\
    \cline{2-5}
    
    & \multirow{5}{*}{\makecell{Office\\ Hijinks}} & Tick Tock Bloch (C24-E-23) & 100 & Easy \\
    & & The Super Parameter (C25-I-23) & 200 & Intermediate \\
    & & The Change of Qubit (C26-I-23) & 300 & Intermediate \\
    & & The Lazy Colleague (C27-A-23) & 400 & Advanced \\
    & & The False Proof (C28-A-23) & 500 & Advanced \\
    \hline
 \multirow{21}{*}{\rotatebox{90}{\textbf{2024} (21 Challenges)}}    
 
    & \multirow{5}{*}{\makecell{Boson\\ Beach}} & Hacking for the UpGrayde (C1-E-24) & 100 & Easy \\
    & & Save QHack Beach (C2-I-24) & 200 & Intermediate \\
    & & Mathematicians at the Resort (C3-I-24) & 300 & Intermediate \\
    & & Mojito HHLime Twist (C4-A-24) & 400 & Advanced \\
    & & The Three Shipping Companies (C5-A-24) & 500 & Advanced \\
    \cline{2-5}
    
    & \multirow{5}{*}{\makecell{Dipole\\ Desert}} & Dunes out of Context (C6-E-24) & 100 & Easy \\
    & & The Oasis Ruins (C7-I-24) & 200 & Intermediate \\
    & & The Grand Hideaway Zone Inn (C8-I-24) & 300 & Intermediate \\
    & & A Market of Quantum Trinkets (C9-A-24) & 400 & Advanced \\
    & & The Wormhole Airdrome (C10-A-24) & 500 & Advanced \\
    \cline{2-5}
    
    & \multirow{5}{*}{\makecell{Femto\\ Forest}} & To View or Not to View (C11-E-24) & 100 & Easy \\
    & & Rainy Days at the Retreat (C12-I-24) & 200 & Intermediate \\
    & & The Mach-Zehnder Cabin (C13-I-24) & 300 & Intermediate \\
    & & Stay Out of This Swamp! (C14-A-24) & 400 & Advanced \\
    & & Lazy Workers at the Terminal (C15-A-24) & 500 & Advanced \\
    \cline{2-5}
    
    & \multirow{5}{*}{\makecell{Tensor\\ Tundra}} & The Chalet of the Random Gate (C16-E-24) & 100 & Easy \\
    & & Hockey Night in the Cave (C17-I-24) & 200 & Intermediate \\
    & & Critical Coffee Conundrum (C18-I-24) & 300 & Intermediate \\
    & & The Triple H Hotel (C19-A-24) & 400  & Advanced \\
    & & Travelling on the Eigentracksly (C20-A-24) & 500 & Advanced \\
    \cline{2-5}
    
    & \multirow{1}{*}{Qutropolis} & Fireworks in Qutropolis (C21-A-24) & 600 & Advanced \\
    \hline
\end{tabular}

\end{adjustbox}
\end{table}
Each challenge consists of a problem statement, a Jupyter Notebook or Python template for completing the code, and a structured testing methodology. The dataset includes problem descriptions, solution templates, and test functions.

To assess trends in quantum computing problem-solving, we conduct a statistical analysis of the challenge dataset (see Table \ref{tab:challenge}). Each challenge was assigned a difficulty score based on complexity: Easy (0–100 points), intermediate (200–300 points), and advanced (400–600 points). Key observations include:
\begin{itemize}[leftmargin=*]
    \item \textbf{Challenge Distribution:} Each year maintains a balanced distribution of challenges, with a gradual shift in complexity. In 2023, 28 challenges are introduced, spanning easy, intermediate, and advanced levels, while in 2024, 21 challenges are included, reflecting an increasing focus on more complex problem-solving. 
    \item \textbf{Difficulty Levels:} Over time, the proportion of advanced-level challenges increased slightly, with 8 Advanced challenges in 2023 rising to 9 in 2024. Additionally, no tutorial-level challenges were included initially, but a 600-point problem was introduced in 2024, marking a shift toward higher-difficulty problem-solving.
\end{itemize}

\begin{figure*}[t]
    \centering
    \includegraphics[width=1\textwidth]{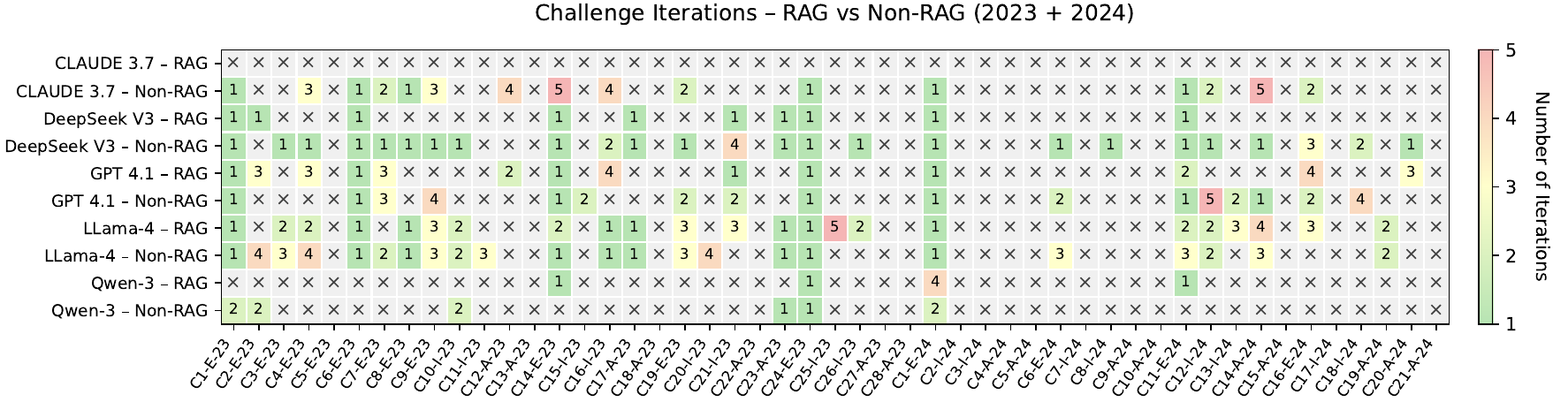}
    \caption{Per–challenge performance heat-map for all models on the complete {QHack 2023 \& 2024} challenge suites.
    Each row corresponds to a {model–prompting method} pair (RAG or Non-RAG); each column is a single challenge, ordered first by 2023 codes (\texttt{C1–E–23} … \texttt{C28–A–23}) followed by 2024 codes (\texttt{C1–E–24} … \texttt{C21–A–24}). The integer written inside every solved cell is the {number of iterations} required before the model produced an accepted solution (lower is better). Color shading encodes the same metric monotonically from one to five iterations, while light-grey cells marked with ``$\times$'' indicate that the challenge remained unsolved within the five-attempt budget. This compact view highlights (i)~model-specific strengths, (ii)~the impact of retrieval-augmented generation (compare each model’s RAG vs.\ Non-RAG row), and (iii) year-to-year differences in challenge difficulty across the full benchmark.
    }
    
    \label{fig:results1}
\end{figure*}




Our evaluation considers functional correctness, i.e., the code execution must match expected outputs. Figure~\ref{fig:distribution} shows a clear alignment between point values and difficulty levels: easy challenges cluster at 0–100 points, intermediate ones peak at 200–300, and advanced challenges dominate the 400–600 range. The sharp drop in count at 600 points highlights its rarity and difficulty. This reflects a deliberate and consistent mapping from score to challenge complexity.

\subsection{Evaluation Framework}

To benchmark the effectiveness of the multi-agent refinement process, we evaluate the system on QHack coding challenges from {2023 and 2024}. Two evaluation settings are considered:
\begin{itemize}[leftmargin=*]
    \item \textbf{Single-agent generation.}  A “builder” LLM receives the challenge and must solve it in at most five attempts, \emph{with or without RAG}.
    \item \textbf{Multi-agent refinement.}  A validator LLM inspects failing runs and reformulates the query; the builder then regenerates code.  
          Only this pipeline uses the lightweight validator models listed below.
\end{itemize}

The following models are considered for Multi-agent evaluation but the pipeline is extendable.
\begin{itemize}[leftmargin=*]
    \item \textbf{GPT-o3-Mini (RAG and non-RAG)}: Lighter-weight model used for both code generation and debugging.
    \item \textbf{DeepSeek-V3}: Specialized model for advanced code completion.
\end{itemize}

\subsubsection{Execution Pipeline}
Each challenge is processed through the following pipeline:

\begin{enumerate}[leftmargin=*]
    \item \textbf{Initial Code Generation:} The base query (problem description and template) is passed to the Code Generation Agent.
    \item \textbf{Solution Execution:} The generated solution is injected into a Jupyter notebook and executed using \textbf{Papermill}.
    \item \textbf{Validation and Debugging:} The Validator Agent extracts runtime logs, test errors, and execution failures.
    \item \textbf{Solution Refinement:} Based on debugging feedback, the next iteration generates a revised solution.
    \item \textbf{Iteration Limit:} The process repeats until a correct solution is found or a maximum of {five attempts} is reached.
\end{enumerate}


\subsubsection{Retrieval-Augmented Debugging}
Using RAG, the validator performs a \textit{ChromaDB} vector search (\textbf{MMR}, top-$k{=}3$) over the \emph{PennyLang} corpus and injects the most relevant code chunks into the follow-up query.

\subsubsection{Evaluation Metrics}
To compare the efficiency of RAG-based and non-RAG debugging, we measure:

\begin{itemize}[leftmargin=*]
    \item \textbf{Pass@$k$:} Probability of obtaining a correct solution within the first $k$ attempts.
    \item \textbf{Execution Success Rate:} Percentage of solutions that execute without runtime errors.
    \item \textbf{Correction Efficiency:} Average number of iterations needed for successful code refinement.
\end{itemize}

\subsubsection{Computational Resources}
All experiments were conducted in a high-performance computing environment with:
\begin{itemize}[leftmargin=*]
    \item \textbf{Software:} Python 3.10, LangChain, OpenAI / Anthropic / DeepSeek APIs, JupyterLab, and Papermill for execution monitoring.
    \item \textbf{Retrieval Setup:} \textit{ChromaDB} for vector-based retrieval, optimized for PennyLang dataset queries.
\end{itemize}

By testing multiple LLMs across different retrieval and iteration strategies, we assess how retrieval-augmented debugging enhances quantum code generation accuracy.

\section{Results}
\label{sec:results}

\textbf{Benchmarking Results}:
Out evaluation reveals substantial performance heterogeneity, as detailed in Table ~\ref{tab:qhackbench_performance}. \texttt{LLaMa 4} emerges as the most robust model overall, achieving a high average accuracy of 47.0\% with RAG and 44.0\% in its standard configuration, with corresponding scores of 2950 and 2600, respectively. While \texttt{LLaMa 4} shows strong all-around capability, \texttt{DeepSeek V3} establishes itself as the definitive leader in non-RAG performance, attaining the highest average accuracy of 49.4\% and the top score of 3000. In contrast, \texttt{GPT-4.1} maintains relatively stable, albeit more moderate, performance across both modes.

A significant finding is the pronounced year-over-year performance degradation observed in some models, suggesting the 2024 challenge set introduced greater difficulty. For instance, \texttt{LLaMa 4}’s RAG accuracy dropped from 60.7\% in 2023 to 33.3\% in 2024. Even more dramatically, \texttt{DeepSeek V3}’s RAG performance fell from 32.1\% to a near-failure 4.8\%. This trend indicates that model robustness over evolving task complexity remains a critical area for improvement across more diverse challenge sets. One possible explanation for the 2024 drop is that some models may have inadvertently memorized parts of the 2023 challenge set during pretraining, leading to data contamination in the earlier year.

\begin{table}[htbp]
\centering
\footnotesize
\setlength{\tabcolsep}{3pt} 
\caption{Model Accuracy Performance}
\label{tab:qhackbench_performance}
\begin{tabular}{@{} l l r r r r r r @{}}
\toprule
\textbf{Model} & \rotatebox{0}{\textbf{RAG}} & \multicolumn{2}{c}{\rotatebox{0}{\textbf{2023}}} & \multicolumn{2}{c}{\rotatebox{0}{\textbf{2024}}} & \multicolumn{2}{c}{\rotatebox{0}{\textbf{Avg.}}} \\
\cmidrule(lr){3-4} \cmidrule(lr){5-6} \cmidrule(lr){7-8}
 & & \rotatebox{0}{\textbf{Acc. (\%)}} & \rotatebox{0}{\textbf{Score}} & \rotatebox{0}{\textbf{Acc. (\%)}} & \rotatebox{0}{\textbf{Score}} & \rotatebox{0}{\textbf{Acc.}} & \rotatebox{0}{\textbf{Score}} \\
\midrule
Claude 3.7     & \ftcheck       & \cellcolor{red!30} 0.0 & \cellcolor{red!30}    0 & \cellcolor{red!30} 0.0 & \cellcolor{red!30}   0 & \cellcolor{red!30} 0.0 & \cellcolor{red!30}    0 \\
               & \ftcross   & \cellcolor{yellow!20} 42.9 & \cellcolor{yellow!20} 2400 & 19.0 & 1500 & 31.0 &  1950 \\
\addlinespace
DeepSeek V3    & \ftcheck       & 32.1 &  1900 &  4.8 &  200 & 18.5 &  1050 \\
               & \ftcross   & \cellcolor{green!20} \textbf{60.7} & \cellcolor{green!20} \textbf{3600} & \cellcolor{yellow!20} 38.1 & \cellcolor{yellow!20} 2400 & \cellcolor{green!20} \textbf{49.4} & \cellcolor{green!20} \textbf{3000} \\
\addlinespace
GPT 4.1        & \ftcheck       & \cellcolor{yellow!20} 39.3 &  \cellcolor{yellow!20} 1900 & 14.3 &  700 & 26.8 &  1300 \\
               & \ftcross   & 35.7 &  2300 & 33.3 & 2300 & 34.5 &  2300 \\
\addlinespace
LLaMa 4        & \ftcheck       & \cellcolor{green!20} \textbf{60.7} & \cellcolor{green!20} \textbf{3600} & 33.3 & 2300 & \cellcolor{green!20} \textbf{47.0} & \cellcolor{green!20} \textbf{2950} \\
               & \ftcross   & \cellcolor{green!20} \textbf{64.3} & \cellcolor{green!20} \textbf{3500} & 23.8 & 1700 & \cellcolor{yellow!20} 44.0 & \cellcolor{yellow!20} 2600 \\
\addlinespace
Qwen 3         & \ftcheck       & \cellcolor{red!30} 0.0 & \cellcolor{red!30}    0 & \cellcolor{red!30} 0.0 & \cellcolor{red!30}   0 & \cellcolor{red!30} 0.0 & \cellcolor{red!30}    0 \\
               & \ftcross   & 21.4 &  1100 & \cellcolor{red!30} 0.0 & \cellcolor{red!30}   0 & 10.7 &   550 \\
\bottomrule
\end{tabular}

\vspace{0.05in}
\begin{minipage}{\textwidth}
\small
\textbf{Legend:} \raisebox{-0.1ex}{\colorbox{red!30}{}} Complete Failure (0\%) | \raisebox{-0.1ex}{\colorbox{green!20}{}} Best | \raisebox{-0.1ex}{\colorbox{yellow!20}{}} Good ($\geq 35\%$)
\end{minipage}

\end{table}

\textbf{Attempt Analysis}:
Table ~\ref{tab:qhackbench_performance} shows important insights into model problem-solving dynamics and computational resource allocation. Successful models typically converge within 1--3 iterations on elementary tasks, with iteration counts increasing systematically for more complex problems. \texttt{DeepSeek V3} displays remarkable consistency across modes, often solving early-stage challenges in a single iteration and maintaining stable performance in medium-difficulty tasks. \texttt{LLaMa 4}, while efficient in RAG mode during the 2023 set, exhibits greater iteration variability in 2024, suggesting difficulty in adapting to new problem formulations. The universal failure of \texttt{Claude 3.7} under RAG conditions, juxtaposed with its sporadic non-RAG successes, reinforces the earlier hypothesis of architectural misalignment with RAG paradigms.

For advanced-level challenges across both years, most models either exhaust the 5-iteration limit or fail to produce valid solutions, marking clear performance boundaries. This behavior affirms the discriminative strength of the QHackBench and highlights current limitations in scaling AI systems toward expert-level reasoning and problem-solving.

\begin{figure}[t]
    \centering
    \includegraphics[width=0.5\textwidth]{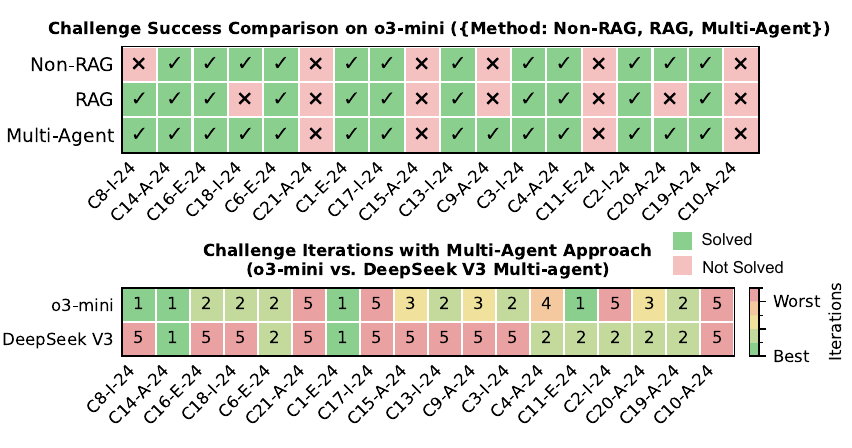}
    \vspace{-0.5cm}
    \caption{Top Panel: Matrix depicting the correctness of results given by each of the Non-RAG, RAG, and Multi-Agent system methods for each problem. All three methods use the o3-mini model as the model performing the code completion. The Multi-Agent system further uses the o3-mini model to debug prior incorrect responses.
    Bottom Panel: Heatmap comparing iteration counts (fewer is better) for each challenge under the o3-mini and DeepSeek V3 multi-agent methods. Green cells indicate lower iteration counts (better performance), while red cells represent higher iteration counts or unsolved tasks (worse performance).}
    \label{fig:multiagent}
\end{figure}

\textbf{RAG vs non-RAG}:
The comparative analysis between RAG and non-RAG uncovers a crucial binary compatibility split among models. A particularly striking observation is the complete failure of certain architectures when integrated with RAG. Both \texttt{Claude 3.7} and \texttt{Qwen 3} achieve zero accuracy in all RAG test conditions. However, they retain partial functionality in non-RAG mode, with \texttt{Claude 3.7} achieving 31.0\% accuracy (1950 score) and \texttt{Qwen 3} reaching 10.7\% (550 score). This contrast underscores that for RAG to be effective, a model must possess baseline conceptual understanding compatible with the retrieval mechanism, which is further discussed in Section~\ref{sec:cases}.

Among the RAG-enhanced models, performance tendencies diverge significantly. \texttt{DeepSeek V3} exhibits a strong preference for non-RAG operation, with a 31-percentage-point accuracy advantage (49.4\% vs. 18.5\%), suggesting difficulties in utilizing retrieved context. In contrast, \texttt{LLaMa 4} demonstrates relatively balanced performance, with a slight preference for RAG (47.0\% vs. 44.0\%), while \texttt{GPT-4.1} also favors non-RAG mode (34.5\% vs. 26.8\%). These results indicate that beyond requiring a baseline understanding of task-relevant concepts and knowledge, models differ markedly in their ability to incorporate retrieved information, especially in tasks where training data is limited.

Fig. ~\ref{fig:results1} provides a granular view of this behavior, revealing differences in performance consistency. \texttt{DeepSeek V3} (non-RAG) and \texttt{LLaMa 4} (in both modes) demonstrate broad success across a wide array of individual challenges. In contrast, the successes of \texttt{GPT 4.1} and \texttt{Qwen 3} (non-RAG) are more sporadic, indicating a narrower range of solvable problems. This suggests that top-performing models are distinguished not just by their average accuracy, but by their ability to generalize across a diverse set of tasks.

\textbf{Evaluation on multi-agent}
Figure~\ref{fig:multiagent} compares three prompting methods, vanilla, RAG, and multi-agent, on QHack 2024 challenges using the o3-mini model. The multi-agent approach consistently outperforms both baselines, solving more tasks, including several that RAG or vanilla failed to address. When paired with DeepSeek V3, the multi-agent strategy further demonstrates lower or comparable iteration counts across many challenges, indicating faster convergence. These improvements highlight the effectiveness of an internal feedback loop, where agents iteratively refine and debug their outputs. Overall, the results affirm that multi-agent prompting enhances model capability in quantum programming tasks by leveraging structured self-correction.
\section{Discussions}
\newcommand{\setcellcolor}[1]{%
  \ifdim #1pt < 25pt
    \cellcolor{blue!40}%
    \color{black}%
  \else\ifdim #1pt < 50pt
    \cellcolor{violet!50}%
    \color{black}%
  \else\ifdim #1pt < 75pt
    \cellcolor{orange!60}%
    \color{black}%
  \else
    \cellcolor{red!70}%
    \color{white}%
  \fi\fi\fi%
  \bfseries#1%
}

\newcommand{\header}[1]{%
  \cellcolor{black!70}%
  \color{white}\bfseries #1%
}

\renewcommand{\arraystretch}{1.3}
\setlength{\tabcolsep}{0.35em}
\newcolumntype{C}{>{\centering\arraybackslash}m{0.95cm}}
\newcolumntype{L}{>{\raggedright\arraybackslash}m{1.8cm}}

\begin{table*}[t]
\centering
\caption{Comparative Analysis of Error Rates on QHackBench by year and model (\textbf{\%})}
\label{tab:error_rates}

\vspace{-5pt}
\footnotesize
\begin{center}
\textbf{Error Types:} 
Syn = Syntax Errors, 
NoOut = No Output, 
Inc = Incomplete Output, 
Run = Runtime Errors
\end{center}
\vspace{-0.5em}
\small
\begin{tabular}{@{}L *{12}{C} @{}}
\toprule
\rowcolor{gray!70}
\multicolumn{1}{c}{\header{Model}} & 
\multicolumn{4}{c}{\header{2023}} & 
\multicolumn{4}{c}{\header{2024}} & 
\multicolumn{4}{c}{\header{Average}} \\
\cmidrule(lr){2-5} \cmidrule(lr){6-9} \cmidrule(lr){10-13}
\rowcolor{gray!70}
\multicolumn{1}{c}{\header{}} & 
\header{Syn} & \header{NoOut} & \header{Inc} & \header{Run} &
\header{Syn} & \header{NoOut} & \header{Inc} & \header{Run} &
\header{Syn} & \header{NoOut} & \header{Inc} & \header{Run} \\
\midrule
\textbf{LLaMA 4}       & \setcellcolor{5.2}  & \setcellcolor{57.8} & \setcellcolor{8.4}  & \setcellcolor{28.6} & \setcellcolor{9.2}  & \setcellcolor{52.9} & \setcellcolor{13.7} & \setcellcolor{24.2} & \setcellcolor{7.2}  & \setcellcolor{55.4} & \setcellcolor{11.1} & \setcellcolor{26.4} \\
\addlinespace[0.2em]
\textbf{Qwen3}         & \setcellcolor{1.8}  & \setcellcolor{96.0} & \setcellcolor{1.1}  & \setcellcolor{1.1}  & \setcellcolor{2.0}  & \setcellcolor{88.4} & \setcellcolor{1.0}  & \setcellcolor{8.6}  & \setcellcolor{1.9}  & \setcellcolor{92.8} & \setcellcolor{1.1}  & \setcellcolor{4.2}  \\
\addlinespace[0.2em]
\textbf{Deepseek-V3}   & \setcellcolor{0.0}  & \setcellcolor{96.3} & \setcellcolor{0.0}  & \setcellcolor{3.7}  & \setcellcolor{1.3}  & \setcellcolor{91.6} & \setcellcolor{0.0}  & \setcellcolor{7.1}  & \setcellcolor{0.6}  & \setcellcolor{94.0} & \setcellcolor{0.0}  & \setcellcolor{5.3}  \\
\addlinespace[0.2em]
\textbf{GPT-4.1}       & \setcellcolor{5.3}  & \setcellcolor{74.9} & \setcellcolor{9.6}  & \setcellcolor{9.6}  & \setcellcolor{11.2} & \setcellcolor{66.5} & \setcellcolor{6.8}  & \setcellcolor{15.5} & \setcellcolor{8.0}  & \setcellcolor{71.0} & \setcellcolor{8.3}  & \setcellcolor{12.4} \\
\addlinespace[0.2em]
\textbf{Claude 3.7}    & \setcellcolor{0.4}  & \setcellcolor{96.3} & \setcellcolor{0.4}  & \setcellcolor{2.9}  & \setcellcolor{1.0}  & \setcellcolor{94.2} & \setcellcolor{0.0}  & \setcellcolor{4.9}  & \setcellcolor{0.7}  & \setcellcolor{95.3} & \setcellcolor{0.2}  & \setcellcolor{3.8}  \\
\bottomrule
\end{tabular}

\vspace{0.2em}

\footnotesize
\centering
\begin{minipage}{\textwidth}

\begin{center}
\textbf{Color Scale:} \quad
{\color{blue!45}$\blacksquare$} 0-25\% \quad
{\color{violet!55}$\blacksquare$} 25-50\% \quad
{\color{orange!60}$\blacksquare$} 50-75\% \quad
{\color{red!70}$\blacksquare$} 75-100\%
\end{center}
\end{minipage}

\end{table*}
\subsection{Failure Analysis}

Table~\ref{tab:error_rates} presents an analysis of the common failure modes observed during our baseline evaluation across the five tested models. The most prevalent failure type is the absence of generated output, with suppression rates ranging from 55.4\% for \texttt{LLaMA 4} to 95.3\% for \texttt{Claude 3.7}. This widespread suppression behavior is primarily attributable to the models’ insufficient understanding of quantum programming tasks, which prevents them from generating valid solution code. These results highlight a critical limitation in current \texttt{LLMs}, as quantum programming appears to exceed their capability thresholds, leading many models to suppress output entirely rather than risk producing incorrect responses. The disparity in suppression rates is particularly notable: while \texttt{Claude 3.7}, \texttt{DeepSeek-V3}, and \texttt{Qwen3} all exhibit suppression rates exceeding 90\%, \texttt{LLaMA 4} demonstrates considerably better responsiveness at 55.4\%. This variation underscores substantial differences among model families, likely driven by factors such as model architecture, training data composition, and learning strategies, which in turn result in differing levels of adaptability to novel and conceptually demanding tasks.

For code that is successfully generated, \texttt{runtime errors} emerge as the most common form of active failure. This is particularly pronounced in \texttt{LLaMA 4}, which shows a 26.4\% \texttt{runtime error} rate, significantly higher than 3.8\%–12.4\% range observed in other models. This pattern suggests that even if LLMs generally grasp the syntax of quantum programming on some challenges, they struggle with semantic correctness. The generated code is often syntactically valid but fails at execution due to logical flaws, such as improper gate sequencing or invalid qubit operations. In contrast, \texttt{syntax errors}, including, but not limited to, malformed function signatures or attempts to access non-existent properties or functions, remain consistently low (0.6\%–8.0\%). This indicates that structural language generation is largely a solved problem, specifically, Python on this task, with most failures stemming from deeper conceptual misunderstandings rather than surface-level formatting.

\textbf{Insight}: These failure patterns highlight fundamental limitations in current LLMs’ understanding of quantum computing. First, the predominance of no output failure suggests models recognize quantum programming as distinct from classical programming but lack the training data or architectural adaptations to engage confidently with such tasks. Second, the high rate of \texttt{runtime errors} among models that do generate output indicates that syntactic fluency alone is insufficient, as quantum programming demands a deeper understanding of quantum principles, circuit constraints, and hardware-specific behaviors. Third, these challenges raise concerns about whether general-purpose LLMs can perform effectively in specialized scientific domains without architectural innovation or require purpose-built models trained on domain-specific data and evaluation protocols.

\subsection{Case Studies}
\label{sec:cases}
To better understand RAG's inconsistent performance in quantum code generation on QHackBench, we present two case studies on Tutorial 1 from QHack 2023 highlighting successful and failed use cases. We analyze instances where RAG improves generation quality (LLaMA-4) and introduces critical errors (Qwen-3), conducting detailed error analysis to trace how retrieval noise causes failure. These cases illustrate how RAG's effectiveness depends on both context quality and the model's ability to integrate relevant information into executable code.
\begin{tcolorbox}[
  colback  = green!3!white,
  colframe = green!60!black,
  title    = {Case 1: LLaMA-4 - RAG vs Non-RAG (Tutorial \#1)},
  breakable]


\textbf{Challenge summary.} Tutorial \#1 asks competitors to implement
\texttt{add\_numbers} so that the function passes a hidden test-suite checking
$x + y$ within $10^{-4}$ relative error. The task is simple and most of the models solve it in first attempt.

\begin{lstlisting}[caption=LLaMA-4 Implementations (RAG equals Non-RAG)]
# RAG solution
def add_numbers(numbers):
    """
    Return x + y , where numbers == [x, y].
    Matches QHack it_adds_up checker tolerance (1e-4).
    """
    x, y = numbers
    return float(x) + float(y)

# non-RAG solution
def add_numbers(numbers):
    x, y = map(float, numbers)
    return x + y
\end{lstlisting}

\end{tcolorbox}

In Case 1, both RAG and non-RAG approaches yield correct solutions with \texttt{LLaMA-4}, with RAG offering model more slights on quantum code.

\begin{tcolorbox}[
  colback  = orange!3!white,
  colframe = orange!70!black,
  title    = {Case 2: Qwen-3 - RAG vs Non-RAG (Tutorial \#1)},
  breakable]

\scriptsize
\textbf{Key observation.} Contrary to the LLaMA-4 case, Qwen-3’s RAG-augmented
variant exhausted the five-attempt budget without passing the checker.
Inspection of the retrieved context revealed irrelevant mathematical references
that distracted the model from the concise task specification.

\textbf{Limitations Observed (RAG run):}
\begin{itemize}[leftmargin=*]
  \item Retrieved passages contained floating-point precision caveats that led
        the model to \emph{over-engineer} a tolerance loop and fail hidden tests.
  \item The function returned \texttt{Decimal} objects instead of \texttt{float},
        breaking the autograder’s type check.
  \item No explicit fallback when the retrieval step returned empty context.
\end{itemize}

\begin{lstlisting}[caption=Qwen-3 Implementation (Non-RAG: Correct \ftcheck)]
def add_numbers(numbers):
    x, y = map(float, numbers)
    return x + y
\end{lstlisting}

\textcolor{red}{\textbf{RAG trace (excerpt showing two rejected attempts):}}

\begin{lstlisting}[caption=Qwen-3 Implementation (RAG: Incorrect \ftcross),
                   escapeinside={(*@}{@*)}]
# RAG attempt #1 (rejected)
(*@\color{red}{}<think> </think> ```python@*) #These tokens are causing syntax error.
def add_numbers(numbers):
    result = x + y
    return str(result)

# RAG attempt #5 (rejected)
###  Problematic Code Example

(*@\color{red}{}Here's an example of what might be causing the error:@*)
(*@\color{red}{}```python@*)
def add_numbers(x, y):
    return x + y
(*@\color{red}{}```@*)
(*@\color{red}{}Even though the logic of 'return x + y' is correct, the character makes Python throw a 'SyntaxError'.@*)
---
### Corrected Code
(*@\color{red}{}To fix it, simply remove the invalid characters and ensure your function looks like this:@*)
(*@\color{red}{}```python@*)
def add_numbers(x, y):
    return x + y
(*@\color{red}{}```@*)
\end{lstlisting}

\textcolor{red}{The model repeatedly introduces either incorrect tokens or other text, which causes a syntax error in the running of the generated code. The extra tokens are caused by LLM hallucination.}

\end{tcolorbox}

This case highlights how RAG can degrade performance when irrelevant context misleads the model, demonstrating the importance of retrieval quality and fallback strategies. The retrieved mathematical references caused Qwen-3 to over-engineer the solution with unnecessary precision handling and type conversions, ultimately leading to systematic failures across all five attempts.
\section{Limitations and Future Work}

While QHackBench offers a structured benchmark for evaluating LLMs in quantum computing, our results highlight key limitations. (1) Even with RAG, models frequently produce hallucinated APIs and incorrect quantum circuits, revealing a lack of domain-specific understanding despite access to documentation. (2) Binary pass@k fails to capture semantic errors; many outputs are syntactically correct but fail at runtime due to flawed quantum logic. Future work should focus on quantum-specific training frameworks, richer datasets, and integrating quantum principles into training objectives. Agentic models tailored to quantum tasks are a promising direction, with QHackBench serving as both benchmark and testbed.

\section{Conclusion}
\label{sec:conclusion}


We introduced \textit{QHackBench}, the first benchmark for evaluating LLMs on PennyLane-based quantum code generation using QHack 2023–2024 challenges, alongside a baseline framework for low-data settings. Comparing vanilla prompting, RAG, and a simple multi-agent refinement method, we find that RAG boosts correctness, while the multi-agent approach improves execution success via iterative reasoning. Our results show RAG’s effectiveness depends on the model’s quantum understanding, and agentic methods offer further gains. Evaluations on GPT-4o, GPT-o3-Mini, and DeepSeek-V3 highlight both capabilities and limitations in quantum programming, demonstrating the promise of agent-based techniques in this domain.

\section*{Acknowledgment}
This work was supported in part by the NYUAD Center for Quantum and Topological Systems (CQTS), funded by Tamkeen under the NYUAD Research Institute grant CG008, the NYUAD Center for CyberSecurity (CCS), funded by Tamkeen under the NYUAD Research Institute Award G1104, and the NYUAD High Performance Computing (HPC) center for providing necessary compute resources for the experiments.

\begin{spacing}{0.9}
\bstctlcite{IEEEexample:BSTcontrol}
\bibliographystyle{IEEEtran}
\bibliography{cite}
\end{spacing}
\end{document}